\documentclass[10pt,twocolumn,letterpaper]{article}

\usepackage[pagenumbers]{cvpr} 

\usepackage{graphicx}
\usepackage{amsmath}
\usepackage{amssymb}
\usepackage{booktabs}
\usepackage{xspace}
\usepackage{interval}
\usepackage{bm}
\usepackage{mathtools}
\usepackage{footmisc}

\usepackage{siunitx}
\usepackage{bbold}
\usepackage{multirow}
\usepackage{bbm}
\usepackage{amssymb}
\usepackage{bbding}
\usepackage[dvipsnames]{xcolor}
\usepackage{makecell}
\usepackage[inline]{enumitem}

\usepackage{algorithm}
\usepackage{listings}

\usepackage{etoolbox}
\makeatletter
\AfterEndEnvironment{algorithm}{\let\@algcomment\relax}
\AtEndEnvironment{algorithm}{\kern2pt\hrule\relax\vskip3pt\@algcomment}
\let\@algcomment\relax
\newcommand\algcomment[1]{\def\@algcomment{\footnotesize#1}}
\renewcommand\fs@ruled{\def\@fs@cfont{\bfseries}\let\@fs@capt\floatc@ruled
  \def\@fs@pre{\hrule height.8pt depth0pt \kern2pt}%
  \def\@fs@post{}%
  \def\@fs@mid{\kern2pt\hrule\kern2pt}%
  \let\@fs@iftopcapt\iftrue}
\makeatother

\usepackage[pagebackref=true,breaklinks=true,letterpaper=true,colorlinks,citecolor=ForestGreen, bookmarks=false]{hyperref}

\newcommand*{\system}{MVD\@\xspace}

\usepackage{pifont}
\newcommand{\cmark}{\ding{51}}
\newcommand{\xmark}{\ding{55}}
\usepackage{caption}
\usepackage{dblfloatfix}

\definecolor{demphcolor}{gray}{.5}

\newcommand{\Up}[1]{\textcolor{ForestGreen}{\xspace\small{\bf $\uparrow$#1}}}
\newcommand{\Down}[1]{\textcolor{red}{\xspace\small{\bf $\downarrow$#1}}}

\newlength\savewidth\newcommand\shline{\noalign{\global\savewidth\arrayrulewidth
\global\arrayrulewidth 1pt}\hline\noalign{\global\arrayrulewidth\savewidth}}
\newcommand{\tablestyle}[2]{\setlength{\tabcolsep}{#1}\renewcommand{\arraystretch}{#2}\centering\small}
\makeatletter\renewcommand\paragraph{\@startsection{paragraph}{4}{\z@}
{.4em \@plus1ex \@minus.2ex}{-.5em}{\normalfont\normalsize\bfseries}}\makeatother

\def\x{$\times$}

\usepackage[capitalize]{cleveref}
\crefname{section}{Sec.}{Secs.}
\Crefname{section}{Section}{Sections}
\Crefname{table}{Table}{Tables}
\crefname{table}{Tab.}{Tabs.}

\def\cvprPaperID{8114} 
\def\confName{CVPR}
\def\confYear{2023}

\begin{document}

\title{Masked Video Distillation: Rethinking Masked Feature Modeling for Self-supervised Video Representation Learning}

\author{Rui Wang$^{1}$\footnotemark[1] \quad Dongdong Chen$^{2}$ \quad Zuxuan Wu$^{1}$\footnotemark[2] \quad Yinpeng Chen$^{2}$ \quad Xiyang Dai$^{2}$\\ \quad Mengchen Liu$^{2}$ \quad Lu Yuan$^{2}$ \quad Yu-Gang Jiang$^{1}$\footnotemark[2] \\
\normalsize$^{1}$Shanghai Key Lab of Intelligent Information Processing, \\ \normalsize School of Computer Science, Fudan Univeristy\\
\normalsize$^{2}$Microsoft Cloud + AI
}
\maketitle

\renewcommand{\thefootnote}{\fnsymbol{footnote}}
\footnotetext[1]{Work done during an internship at Microsoft} \footnotetext[2]{Corresponding authors}

\begin{abstract}
Benefiting from masked visual modeling, self-supervised video representation learning has achieved remarkable progress. However, existing methods focus on learning representations from scratch through reconstructing low-level features like raw pixel RGB values. In this paper, we propose masked video distillation (MVD), a simple yet effective two-stage masked feature modeling framework for video representation learning: firstly we pretrain an image (or video) model by recovering low-level features of masked patches, then we use the resulting features as targets for masked feature modeling. For the choice of teacher models, we observe that students taught by \textbf{video teachers} perform better on temporally-heavy video tasks, while \textbf{image teachers} transfer stronger spatial representations for spatially-heavy video tasks. Visualization analysis also indicates different teachers produce different learned patterns for students.
Motivated by this observation, to leverage the advantage of different teachers, we design a spatial-temporal co-teaching method for MVD. Specifically, we distill student models from both video teachers and image teachers by masked feature modeling. Extensive experimental results demonstrate that video transformers pretrained with spatial-temporal co-teaching outperform models distilled with a single teacher on a multitude of video datasets. Our MVD with vanilla ViT achieves state-of-the-art performance compared with previous supervised or self-supervised methods on several challenging video downstream tasks. For example, with the ViT-Large model, our MVD achieves 86.4\%  and 76.7\% Top-1 accuracy on Kinetics-400 and Something-Something-v2, outperforming VideoMAE by 1.2\% and 2.4\% respectively. When a larger ViT-Huge model is adopted, MVD achieves the state-of-the-art performance with \textbf{77.3\%} Top-1 accuracy on Something-Something-v2 and \textbf{41.1} mAP on AVA v2.2. Code will be available at \url{https://github.com/ruiwang2021/mvd}.

\end{abstract}

\section{Introduction}
\label{sec:intro}

\begin{figure}
    \centering
    \includegraphics[width=0.98\linewidth]{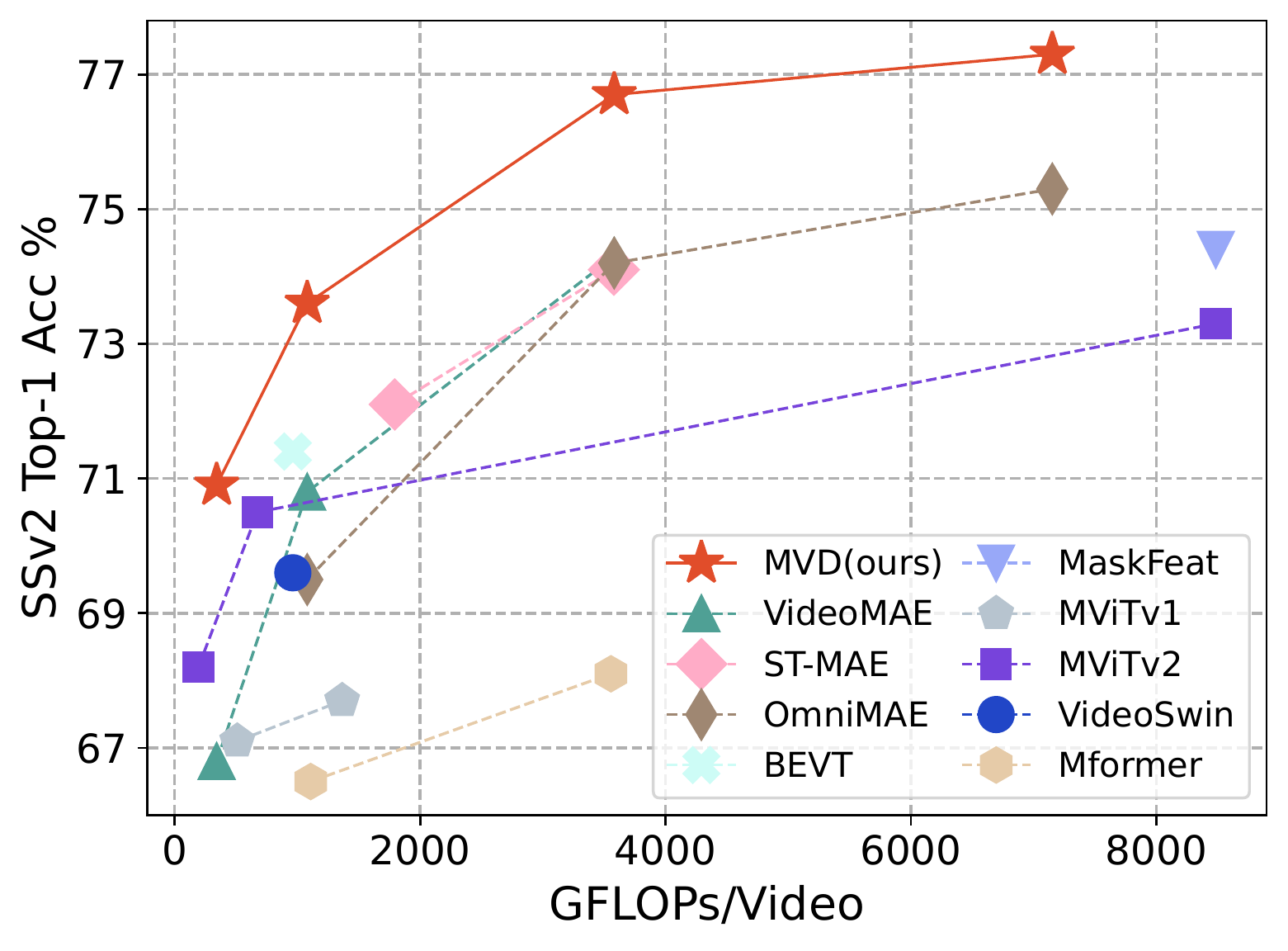}
    \vspace{-0.5em}
    \caption{\textbf{Comparisons of MVD with previous supervised or self-supervised methods on Something-Something v2.} Each line represents the corresponding model of different sizes.}%
    \label{fig:flops_acc_compare}%
\end{figure}

For self-supervised visual representation learning, recent masked image modeling (MIM) methods like MAE~\cite{he2021masked}, BEiT~\cite{bao2021beit} and PeCo \cite{dong2021peco} achieve promising results with vision transformers~\cite{vit} on various vision downstream tasks. Such a pretraining paradigm has also been adapted to the video domain and boosts video transformers by clear margins compared with supervised pretraining on several video downstream tasks. Representative masked video modeling (MVM) works include BEVT \cite{wang2022bevt}, VideoMAE~\cite{tong2022videomae} and ST-MAE~\cite{feichtenhofer2022masked}.

Following MAE~\cite{he2021masked} and BEiT~\cite{bao2021beit},  existing masked video modeling methods \cite{feichtenhofer2022masked,wang2022bevt,tong2022videomae} pretrain video transformers through reconstructing low-level features, \eg, raw pixel values or low-level VQVAE tokens. However, using low-level features as reconstruction targets often incur much noise. And due to the high redundancy in video data, it is easy for masked video modeling to learn shortcuts, thus resulting in limited transfer performance on downstream tasks. To alleviate this issue, masked video modeling \cite{tong2022videomae} often uses larger masking ratios. 

In this paper, we observe that much better performance on video downstream tasks can be achieved by conducting masked feature prediction by using the high-level features of pretrained MIM and MVM models as masked prediction targets. This can be viewed as two-stage masked video modeling, where MIM pretrained image models (\ie, an image teacher) or MVM pretrained video models (\ie, an video teacher) are obtained in the first stage, and they further act as teachers in the second stage for the student model via providing the high-level feature targets. Therefore, we call this method Masked Video Distillation (MVD). 

More interestingly, we find that student models distilled with different teachers in MVD exhibit different properties on different video downstream tasks. Specifically, students distilled from the image teacher perform better on video tasks that mainly rely on spatial clues, while students distilled from the video teacher model perform better on the video downstream tasks where temporal dynamics are more necessary. We think during the pretraining process of masked video modeling in the first stage, video teachers have learned spatial-temporal context in their high-level features. Therefore, when employing such high-level representations as prediction targets of masked feature modeling, it will help encouraging the student model to learn stronger temporal dynamics. By analogy, image teachers provide high-level features as targets that include more spatial information, which can help the student model learn more spatially meaningful representations.
We further analyze the feature targets provided by image teachers and video teachers, and calculate the cross-frame feature similarity. It shows that the features provided by the video teachers contain more temporal dynamics.

Motivated by the above observation, to leverage the advantages of video teachers and image teachers, we propose a simple yet effective spatial-temporal co-teaching strategy for MVD. In detail, the student model is designed to reconstruct the features coming from both the image teacher and video teacher with two different decoders, so as to learn stronger spatial representation and temporal dynamics at the same time. Experiments demonstrate that MVD with co-teaching from both the image teacher and the video teacher significantly outperforms MVD only using one single teacher on several challenging downstream tasks.

Despite the simplicity, our MVD co-teaching is super effective and achieves very strong performance on multiple standard video recognition benchmarks. For example, on Kinectics-400 and Something-Something-v2 datasets, compared to the baseline without MVD,  MVD co-teaching with 400 epochs using a teacher model of the same size achieves \textbf{ 1.2\%}, \textbf{2.8\%} Top-1 accuracy gain on ViT-B. If a larger teacher model ViT-L is used, more significant performance gains (\ie, \textbf{1.9\%}, \textbf{4.0\%}) can be obtained. When ViT-Large is the target student model, our method can achieves \textbf{86.4\%} and \textbf{76.7\%} Top-1 accuracy on these two datasets, surpassing existing state-of-the-art method VideoMAE \cite{tong2022videomae} by \textbf{1.2\%} and \textbf{2.4\%} respectively. When a larger ViT-Huge model is adopted, MVD achieves the state-of-the-art performance with \textbf{77.3\%} Top-1 accuracy on Something-Something-v2 and \textbf{41.1} mAP on AVA v2.2.

Our contributions can be summarized as below:
\begin{itemize}
    \item We find that using MIM pretrained image models and MVM pretrained video models as teachers to provide the high-level features for continued masked feature prediction can learn better video representation. And representations learned with image teachers and video teachers show different properties on different downstream video datasets.
    \item We propose masked video distillation together with a simple yet effective co-teaching strategy, which enjoys the synergy of image and video teachers.
    \item We demonstrate strong performance on multiple standard video recognition benchmarks, surpassing both the baseline without MVD and prior state-of-the-art methods by clear margins.
\end{itemize}

\section{Related Work}
\label{sec:related_work}

\vspace{0.05in}
\noindent \textbf{Vision transformers for video understanding.} For video understanding tasks, modeling the spatial-temporal information is the most important factor to consider in the architecture design. In the early works of video understanding, common video architectures, e.g., 3D CNNs~\cite{quovadis,c3d,r21d,slowfast,x3d} and 2D CNNs with temporal modules~\cite{recurrentdonahue,recurrentjoe,twostream,tsn,tsm}, are designed by extending existing 2D CNN models on the temporal dimension. Recently, Vision Transformers~\cite{vit,liu2021swin,dong2022cswin} achieve significant progress on several computer vision tasks. Some works also adapt vision transformers to the video domain and achieve superior performance compared to previous CNN-based architectures. For example, TimeSformer~\cite{timesformer} and ViViT~\cite{arnab2021vivit}   study several variants of space-time factorization for extending the plain ViT architecture to video domain. Some works~\cite{xvit,tokenlearner,patrick2021keeping} further explore how to reduce computational cost of the space-time attention. VideoSwin~\cite{liu2021video} and MViT~\cite{fan2021multiscale,mvitv2} study the hierarchical architecture and introduce an inductive locality bias into video transformers. Uniformer~\cite{li2022uniformer} and Video Mobile-Former \cite{wang2022video} propose to integrate 3D CNNs and spatial-temporal self-attention mechanism for efficiency consideration. For convincing performance on the video understanding tasks, most video transformers require model weights pretrained on the large-scale image datasets. In this paper, we explore the self-supervised pretraining of video transformers and show pretraining strategy will significantly influence the downstream performance, which is orthogonal to the transformer architecture design.

\vspace{0.05in}
\noindent\textbf{Self-supervised video representation learning.} The early works \cite{wang2015unsupervised,misra2016shuffle,xu2019self,benaim2020speednet} of self-supervised video representation learning focus on designing the pretext tasks based on the temporal structure of videos. More recently, contrastive learning \cite{chen2020simple,he2020momentum,li2021improve} that forces different views of the same image sample to be closer in the feature space while pushing the views of different images farther becomes a new paradigm of representation learning, and some works \cite{ge2021revitalizing,han2020memory,pan2021videomoco,guo2022cross,cvrl,diba2021vi2clr,feichtenhofer2021large} design the contrastive learning methods on the video domain by exploring effective ways of spatial-temporal augmentations. However, as the learning supervision based on contrastive learning is applied on global representation, it cannot well model the local relationship or learn fine-grained local representation.

\vspace{0.05in}
\noindent\textbf{Masked visual modeling.} Masked language modeling \cite{liu2019roberta,devlin2018bert} has been one of the dominant pretraining methods of language transformers. With the success of vision transformers, masked visual modeling \cite{bao2021beit} has been introduced to self-supervised visual pretraining and demonstrates to be also helpful to multimodal visual-language learning \cite{dong2022maskclip,zheng2022general}. Following BERT~\cite{devlin2018bert} pretraining, BEiT~\cite{bao2021beit} and PeCo~\cite{dong2021peco} pretrain ViT by predicting the discrete visual tokens of masked patches, which are encoded by a pretrained VQ-VAE. MAE~\cite{he2021masked} proposes an asymmetric encoder-decoder framework for the reconstruction of pixels, which reduces the computational cost of masked image modeling significantly. SimMIM~\cite{xie2022simmim} and MaskFeat~\cite{maskfeat} propose to recover low-level features of masked patches like pixels or HOG features with hierarchical ViT. In contrast, iBOT~\cite{zhou2021ibot}, BootMAE~\cite{bootmae} and sdAE~\cite{chen2022sdae} adopt an exponential moving average of the student model as the online teacher model, which makes the target features bootstrapped during training. In the video domain, some pioneering works~\cite{tan2021vimpac,wang2022bevt,tong2022videomae,feichtenhofer2022masked} extend masked image modeling to masked video modeling. BEVT~\cite{wang2022bevt} proposes a two-stream pretraining joint pretraining framework by predicting the discrete tokens with both image transformer and video transformer. VideoMAE~\cite{tong2022videomae} and ST-MAE~\cite{feichtenhofer2022masked} follow MAE and reconstruct the pixels of masked video patches with an extremely high masking ratio. Unlike most previous works of masked video modeling, our MVD focuses on masked feature modeling with high-level features as targets, and finds that student models using image and video teacher models will have different properties and complement each other.

\vspace{0.05in}
\noindent\textbf{Knowledge distillation.} Knowledge distillation \cite{hinton2015distilling,phuong2019towards,gou2021knowledge} aims to transfer the knowledge of the teacher model to the student model by adopting the output of the teacher model as the target for training the student model. Typical knowledge distillation works \cite{hinton2015distilling,shen2021fast,shen2020meal} mainly focus on supervised learning, e.g., image classification. Recently, self-supervised knowledge distillation \cite{xu2020knowledge,fang2021seed,xu2021bag} has also been studied  to learn representations from self-supervised pretrained models. In this paper, we present the first attempt that uses the masked image modeling pretrained image and video model as the masked feature prediction target in the video domain. It shows self-supervised MIM pretrained model can further boostrap the mask video pretraining and bring significant performance gain.

 \begin{figure*}[t]
\begin{center}
   \includegraphics[width=0.9\linewidth]{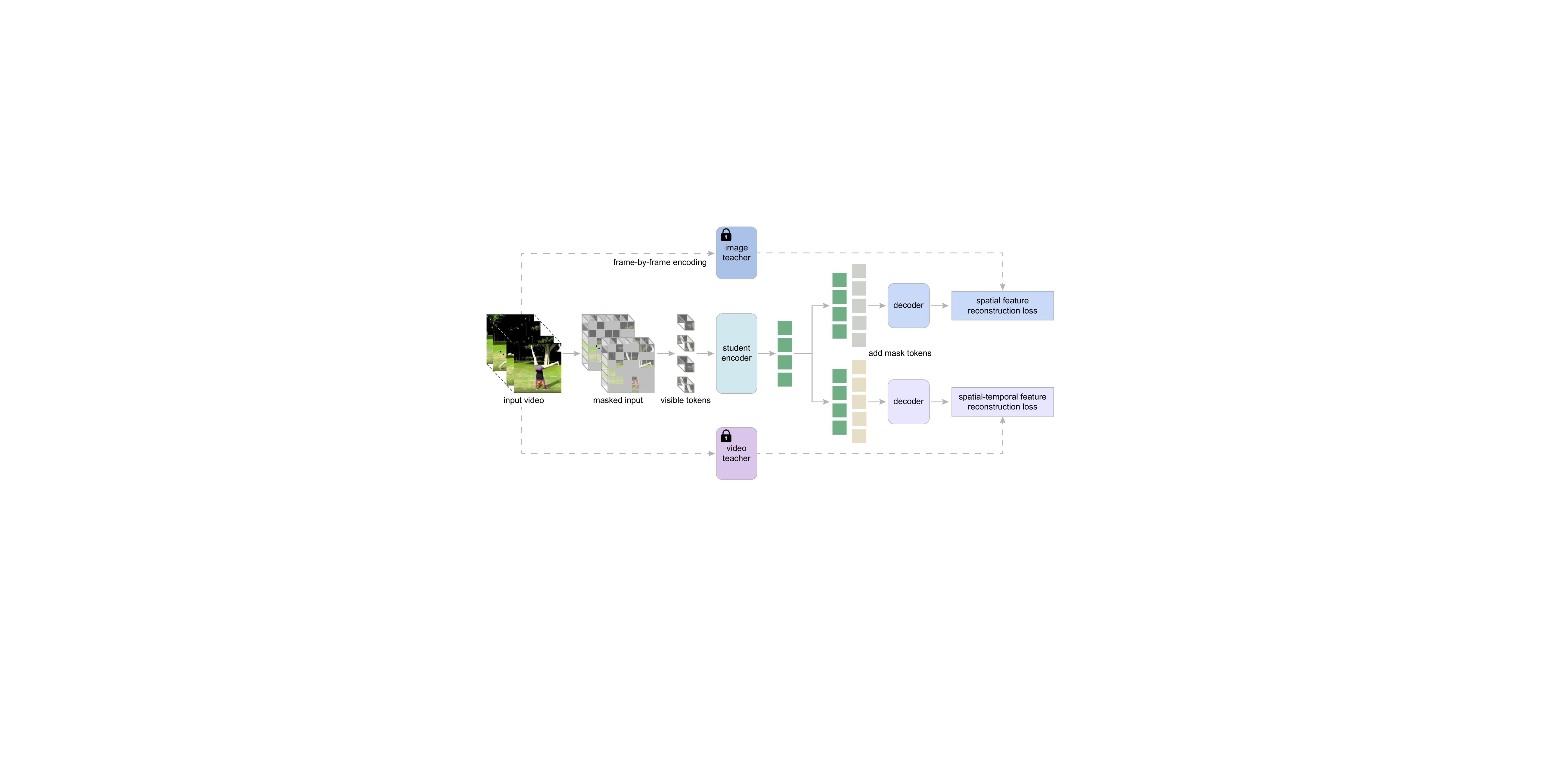}
\end{center}
   \vspace{-0.2in}
   \caption{\textbf{An overview of MVD framework.} Firstly the image teacher is pretrained by masked image modeling and the video teacher is pretrained by masked video modeling. Then the student model is trained from scratch to predict target high-level features encoded by the image teacher and the video teacher. The teacher models are fixed in the distillation stage.}
\label{fig:framework}
\vspace{-10pt}
\end{figure*}

\section{Method}
While masked video modeling has demonstrated promising performance for self-supervised learning, most existing approaches reconstruct relatively low-level information in the forms of raw pixels~\cite{tong2022videomae}, low-level features like HOG~\cite{maskfeat} and VQVAE tokens \cite{wang2022bevt}. In this paper, instead of reconstructing low-level information, we conduct masked video modeling at the feature-level. This is achieved by a two-stage framework, \system, optimized to predict high-level features that derived from off-the-shelf MIM pretrained image models \cite{he2021masked} and MVM pretrained video models \cite{tong2022videomae} which are readily available. Below, we first given an overview of the masked feature modeling paradigm in \cref{sec:mfm} and then we introduce our proposed \system in \cref{sec:mvd}. Finally, we present the architectural design of \system in \cref{sec:network}.

\subsection{The Paradigm of Masked Feature Modeling}
\label{sec:mfm}
The core of masked feature modeling is to train models to predict features of masked input regions. In this paper, we follow the decoupled encoder-decoder transformer architecture in MAE~\cite{he2021masked} due to its effectiveness and simplicity. An input $X$ (image $X_{img} \in \mathbb{R}^{H \times W \times 3}$ or video $X_{vid} \in \mathbb{R}^{T \times H \times W \times 3}$) is partitioned into several non-overlapping patches, and then each patch is mapped to a visual token with a linear projection layer. Before feeding tokens to the transformer encoder $f$, a subset of tokens is masked and dropped from the token sequence. To reconstruct the information of masked tokens, the token sequence consisting of the visible tokens from the encoder and learnable mask tokens are input to a shallow transformer decoder $g$:

\begin{equation}
  Y = g(\texttt{concat}(f(X_{vis}), T_m)),
  \label{eq:decoder_output}
\end{equation}

where $X_{vis}$ denotes the visible input tokens, and $T_m$ denotes the mask tokens. The subset of output tokens from decoder corresponded to input mask tokens contains reconstructed information of masked tokens. The reconstruction target for each masked patch $X(p)$ is represented as a patch feature $h(X(p))$.
Here, $h$ is represents a function for generating the target features, \eg $h$ produces low-level RGB values of pixels in the patch in~\cite{tong2022videomae,feichtenhofer2022masked}. Then, to train the encoder and the decoder, a loss function that measures the distance $D$ between the ground-truth features of masked patches and reconstructed ones is defined as:

\begin{equation}
  L_{mfm}(h) = \frac{1}{\lvert M \rvert} \sum_{p \in M} D(Y(p), h(X(p)))
  \label{eq:mfm_loss}
\end{equation}

where $p$ is the token index and $M$ is the set of masked tokens. For pixel regression in MAE~\cite{he2021masked} and VideoMAE~\cite{tong2022videomae}, the L2 distance is used as the distance metric.

\subsection{Masked Video Distillation}
\label{sec:mvd}
In this paper, we propose Masked Video Distillation (MVD), which performs masked feature modeling on videos using high-level features as opposed to low-level pixels. In particular, we simply use outputs generated by off-the-shelf self-supervised pretrained image or video models, which are readily available, as reconstruction targets. These high-level features, serving as targets of the mask \& prediction tasks, are encoded by teacher models pretrained by masked visual modeling like MAE or VideoMAE. For video representation learning, the reconstruction targets can take the form of spatial features encoded by image teacher models, or spatial-temporal features encoded by video teacher models. 
More specifically, the image teachers is pretrained by masked image modeling, while the video teacher is pretrained with masked video modeling, both of which aim at reconstructing raw pixels. Once trained, we use the image encoder $h_{img}$ to generate the spatial targets, and the pretrained video transformer encoder $h_{vid}$ to generate spatial-temporal targets. The loss function of MVD with the image teacher and video teacher can be denoted by $L_{mfm}(h_{img})$ and  $L_{mfm}(h_{vid})$, respectively

\vspace{0.05in}
\noindent \textbf{Spatial-temporal Co-teaching.} When performing MVD with a single teacher, we observe that students distilled from different teachers learn different video representations and perform well on different kinds of downstream video tasks. To improve the performance of MVD on different downstream video tasks, we propose spatial-temporal co-teaching that explores information from both image and video teachers such that the student model can handle videos of different types better. For instance, videos with fastly changing human actions require more temporal information while spatial clues might be sufficient for relatively static videos. To this end, \system is trained to predict target high-level features produced by the image teacher and the video teacher at the same time.
This is achieved by using two separated decoders to reconstruct different target features. The final loss of MVD with spatial-temporal co-teaching is:

\begin{equation}
  L_{mvd} = \lambda_1 L_{mfm}(h_{img}) + \lambda_2 L_{mfm}(h_{vid})
  \label{eq:coteach}
\end{equation}

where $\lambda_1$ and $\lambda_2$ denote the hyper-parameters that balance the weights of the image teacher and the video teacher. The pseudo code of \system is shown in \Cref{alg:code}.

\subsection{Architectural Design}
\label{sec:network}
\noindent \textbf{Encoder.} For MVD, the vanilla transformer backbone is used as the encoder. For a video input $X_{vid} \in \mathbb{R}^{T \times H \times W \times 3}$, we adopt 3D patch embedding with a patch size of $2 \times 16 \times 16$. After patch partitioning and linear embedding, we obtain $T/2 \times H/16 \times W/16$ tokens. For the masked feature modeling task, the tokens are masked with a high masking ratio and the remaining tokens are fed into the transformer layers. For finetuning on downstream tasks, we input all tokens to the subsequent layers. In each layer, joint spatial-temporal self-attention is applied on the whole input token sequence. 

\begin{algorithm}[t]
    \caption{Pseudocode of MVD in PyTorch style.}
    \label{alg:code}
    \algcomment{\fontsize{7.2pt}{0em}\selectfont
    \vspace{-2.em}
    }
    \definecolor{codeblue}{rgb}{0.580,0.337,0.447}
    \lstset{
     backgroundcolor=\color{white},
     basicstyle=\fontsize{7.2pt}{7.2pt}\ttfamily\selectfont,
     columns=fullflexible,
     breaklines=true,
     captionpos=b,
     commentstyle=\fontsize{7.2pt}{7.2pt}\color{codeblue},
     keywordstyle=\fontsize{7.2pt}{7.2pt},
    }

\begin{lstlisting}[language=python, mathescape=true]
# f: student encoder
# g_img: decoder for reconstructing spatial features
# g_vid: decoder for reconstructing spatial-temporal features
# t_m: learnable mask tokens
# h_img: image teacher model
# h_vid: video teacher model

for x, m in loader:  #  x: video data, m: mask 
    x_pe = patch_emb(x)  #  patch embedding of input
    x_vis = mask_select(x_pe, 1 - m)  #  masking tokens
    q_vis = f(x_vis)  # visible local patch features
    
    # reconstruction of target features
    p_img = g_img(concat(q_vis, t_m))
    p_vid = g_vid(concat(q_vis, t_m))

    # compute target features with teacher models
    k_img = h_img(x)  #  target spatial features
    k_vid = h_vid(x)  #  target spatial-temporal features
   
    # compute reconstruction loss
    loss_img = smooth_L1_loss(p_img $\odot$ m, k_img $\odot$ m)
    loss_vid = smooth_L1_loss(p_vid $\odot$ m, k_vid $\odot$ m)

    loss = $\lambda_1$ * loss_img + $\lambda_2$ * loss_vid
    loss.backward()
    optimizer.step()  #  optimizer update


\end{lstlisting}
\end{algorithm}

\vspace{0.05in}
\noindent \textbf{Mask strategy.} For MVD, we follow ~\cite{tong2022videomae} and adopt tube masking for masked feature modeling. First a 2D random mask is generated and then extended along the temporal dimension. Therefore, the spatial mask on each time slice is the same, which prevents information leakage between frames. Tube masking with a high masking ratio (\eg, 90\%) encourages the video transformer to model the high-level semantics during pretraining.

\vspace{0.05in}
\noindent \textbf{Decoder.} For MVD, shallow decoders consist of vanilla transformer layers and a linear projection layer. The transformer layers in decoders are the same as those in the encoder. Since spatial-temporal co-teaching introduces two different reconstruction targets for masked feature modeling, two separated decoders that share the same architecture but contain different  weights are placed on the top of the encoder. Learnable masked tokens corresponded to masked patches are concatenated with visible tokens from the encoder before fed into the decoder. After jointly modeling the spatial-temporal relationships, the output tokens of transformer layers are mapped to final predictions by the linear projection layer.

\vspace{0.05in}
\noindent \textbf{Reconstruction targets.} For generating spatial-temporal target features,  the video teacher, which shares the same architecture as the student model, is pretrained by a VideoMAE \cite{tong2022videomae} manner on the video dataset. For obtaining spatial targets, we adopt the vanilla image ViT pretrained by masked image modeling \cite{he2021masked} on the image dataset (\eg, ImageNet-1K). It is worth noting that one 3D patch (with size of $2 \times 16 \times 16$) for the video transformer  corresponds to two 2D patches (with size of $16 \times 16$) for the image transformer. Following ~\cite{feichtenhofer2022masked}, we predict the spatial features of a single time slice (that is the front 2D patch), which reduce the prediction layer's size. 

\section{Experiments}
In this section, we first introduce the experimental setup in \cref{sec:setup}, and then present the main results in ~\cref{sec:main_results}, followed by an extensive analysis to verify the effectiveness of different components in  \cref{sec:ablation}.

\subsection{Experimental Setup}
\label{sec:setup}
\noindent \textbf{Dataset.} We pretrain the vanilla ViT with MVD on Kinetics-400 by default, and evaluate the learned model on four video recognition downstream tasks: (a) Kinetics-400 (K400)~\cite{quovadis}, which consists of ${\sim}240K$ training videos and ${\sim}20K$ validation videos with an average duration of 10 seconds. All video clips are labeled into 400 classes. (b) Something-Something V2 (SSv2)~\cite{ssv2}, which contains ${\sim}160K$ videos for training and ${\sim}20K$ videos for validation. The videos in SSv2 with an average duration of 4 seconds are labeled into 174 motion-centric categories. (c) UCF-101~\cite{ucf101} is a relatively small dataset, consisting of ${\sim}9.5K$ training videos and ${\sim}3.5K$ validation videos. (d) HMDB51~\cite{kuehne2011hmdb} is also a small video dataset that contains around 3.5K/1.5K train/val videos. On UCF101 and HMDB51, we follow the commonly used protocols and evaluate our method across all 3 train/val splits. We also transfer pretrained models to a challenging spatial-temporal action detection dataset AVA~\cite{gu2018ava}.

\vspace{0.05in}
\noindent \textbf{Implementation details.} Our MVD is performed on vanilla ViTs with different capacities (\ie, ViT-S, ViT-B, ViT-L). By default, image teacher models are pretrained on ImageNet-1K for 1600 epochs and video teachers are pretrained on K400 for 1600 epochs. We follow the training strategy in MAE~\cite{he2021masked} and VideoMAE~\cite{feichtenhofer2022masked} for image teachers and video teachers respectively. In the distillation stage, student models are first pretrained from scratch on K400 for 400 epochs unless mentioned otherwise. The resulting models are then finetuned on downstream video tasks. The video clip length is 16 for both pretraining and finetuning. We adopt AdamW optimizer~\cite{loshchilov2018decoupled} and Smooth L1 loss for the optimization of student models. We conduct experiments of pretraining on 32 NVIDIA V100 GPUs and experiments of finetuning on 16 NVIDIA V100 GPUs. More details are included in supplementary materials.

\begin{table}[t]
\centering
\begin{tabular}{l|c|c|c|c|c}
\multicolumn{1}{l|}{\multirow{2}{*}{student}} & \multicolumn{2}{c|}{teachers} & \multicolumn{1}{l|}{\multirow{2}{*}{K400 top-1}} & \multicolumn{1}{c|}{\multirow{2}{*}{SSv2 top-1}}            \\ 
\cline{2-3} \multicolumn{1}{l|}{} & \multicolumn{1}{l|}{image} & \multicolumn{1}{l|}{video} & \multicolumn{1}{l|}{} & \multicolumn{1}{l|}{}      \\ 
\shline
 & \cmark & \xmark & 80.4  & 69.4 \\
ViT-S & \xmark & \cmark & 80.1 & 70.0 \\
 & \cmark & \cmark & \textbf{80.6}  & \textbf{70.7} \\
\hline
 & \cmark & \xmark & 82.3 & 71.4 \\
ViT-B & \xmark & \cmark & 82.1 & 71.8 \\
 & \cmark & \cmark & \textbf{82.7} & \textbf{72.5} \\
\end{tabular}
\vspace{-2pt}
\caption{\textbf{Co-teaching with the image teacher and the video teacher outperforms distillation with a single teacher in \system.} We adopt ViT-B as the architecture of teachers and distill students for 400 epochs here.}
\label{tab:diff_teacher}
\vspace{-5pt}
\end{table}


\begin{table}[t]
\centering
\tablestyle{5pt}{1.05}
\begin{tabular}{lc|c|c|c|c}
\multicolumn{1}{l}{\multirow{2}{*}{student}} & \multicolumn{1}{l|}{\multirow{2}{*}{teacher}} & \multicolumn{2}{c|}{K400 top-1} & \multicolumn{2}{c}{SSv2 top-1}            \\ \cline{3-6} 
\multicolumn{1}{l}{}                               & \multicolumn{1}{l|}{}                               & \multicolumn{1}{l|}{ViMAE} & MVD    & \multicolumn{1}{l|}{ViMAE} & MVD        \\ 
\shline
ViT-S & ViT-B & 79.0 & 80.6\Up{1.6} & 66.4 & 70.7\Up{4.3} \\
ViT-S & ViT-L &  79.0 & 81.0\Up{2.0} & 66.4 & 70.9\Up{4.5}   \\
\hline
ViT-B & ViT-B & 81.5 & 82.7\Up{1.2} & 69.7 & 72.5\Up{2.8} \\
ViT-B & ViT-L & 81.5 & 83.4\Up{1.9} & 69.7 & 73.7\Up{4.0} \\
\hline
ViT-L & ViT-L & 85.2 & 86.0\Up{0.8} & 74.0 & 76.1\Up{2.1} \\
\end{tabular}
\caption{\textbf{MVD achieves significant improvement compared with VideoMAE across different model scales.} MVD is pretrained on K400 for 400 epochs, while VideoMAE is pretrained on K400 for 1600 epochs here.}
\label{tab:diff_size}
\vspace{-5pt}
\end{table}

\subsection{Main Results}
\label{sec:main_results}
\noindent \textbf{Students distilled from different teachers.} Unlike masked image modeling, masked feature modeling on video data has more choices on  reconstruction targets. Besides spatial features, to include temporal dynamics in the reconstruction target, we can also adopt spatial-temporal features encoded by pretrained video models. In Table~\ref{tab:diff_teacher}, we compare students distilled by the image teacher and the video teacher on K400, a downstream task that mainly relies on spatial clues, and SSv2, a temporally-heavy downstream task. Our observations are as follow: (a) Masked feature modeling with high-level features as targets achieves convincing performance on downstream video tasks, and outperforms VideoMAE baseline significantly (compared with the baseline results in Table~\ref{tab:diff_size}). In particular, with both image and video teachers using a ViT-S as the backbone, \system achieves consistent gains over VideoMAE on both K400 (80.6\% \vs 79.0\%) and SSV2 (70.7\% \vs 66.4\%). 
(b) Students distilled from the image teacher achieve higher top-1 accuracy on K400, while students distilled from the video teacher perform better on SSv2.
For example,  ViT-S achieves an accuracy of 80.4\% and 69.4\% using an image teacher on K400 and SSv2 respectively. With a video teacher, on the other hand, ViT-S offers a top-1 accuracy of 80.1\% and 70.0\% respectively. As it has been shown that videos in K400 less sensitive to temporal modeling compared to SSv2,
the results demonstrate students learn stronger spatial representation from the image teacher while the video teacher transfer more knowledge about temporal dynamics to students. 

\vspace{0.05in}
\noindent \textbf{Co-teaching outperforms distilling with a single teacher.} To improve the performance on different kinds of downstream video tasks, we introduce spatial-temporal co-teaching in MVD, which trains the model to predict spatial features and spatial-temporal features of masked patches in a decoupled way. The results in Table~\ref{tab:diff_teacher} indicate that students distilled with spatial-temporal co-teaching outperforms students distilled from either single teacher on both spatially-heavy task and temporally-heavy task.

\begin{table}[t!]
\centering
\captionsetup[sub]{font=normalsize}
\tablestyle{2.0pt}{1.02}
\resizebox{1.02\linewidth}{!}{
\hspace*{-7pt}
\begin{tabular}{l|l|c|c|c|c}
    method & extra data & top-1 & top-5 & \footnotesize{GFLOPs} & \footnotesize{Param} \\
    \shline \hline
    \textit{supervised} & & & & & \\
    NL I3D R101~\cite{nonlocal} &\multicolumn{1}{c|}{-} & 77.3 & 93.3 & 359\x30 & 62 \\
    ip-CSN-152~\cite{ipcsn} &\multicolumn{1}{c|}{-} & 77.8 & 92.8 & 109\x30 & 33 \\
    {SlowFast} NL~\cite{slowfast} &\multicolumn{1}{c|}{-} & 79.8 & 93.9 & 234\x30 & 60 \\
    X3D-XL~\cite{x3d} &\multicolumn{1}{c|}{-} & 79.1 & 93.9 & 48\x30 & 11 \\ 
    \hline
    MViTv1-B~\cite{fan2021multiscale} &\multicolumn{1}{c|}{-} & 80.2 & 94.4 & 170\x5 & 37 \\
    VideoSwin-B~\cite{liu2021video} & IN-1K & 80.6 & 94.6 & 282\x12 & 88 \\
    Uniformer-B~\cite{li2022uniformer} & IN-1K & 83.0 & 95.4 & 259\x12 & 50 \\
    \hline
    TimeSformer~\cite{timesformer} & IN-21K & 80.7 & 94.7 & 2380\x3 & 121 \\
    Mformer-B~\cite{patrick2021keeping} & IN-21K & 79.7 & 94.2 & 370\x30 & 109 \\
    Mformer-L~\cite{patrick2021keeping} & IN-21K & 80.2 & 94.8 & 1185\x30 & 382 \\
    ViViT-L FE~\cite{arnab2021vivit} & IN-21K & 81.7 & 93.8 & 3980\x3 & N/A \\
    VideoSwin-L ~\cite{liu2021video} & IN-21K & 83.1 & 95.9 & 604\x12 & 197 \\
    \shline\hline
    \textit{self-supervised} & & & & & \\
    VIMPAC {\scriptsize{ViT-L}}~\cite{tan2021vimpac} & HowTo100M & 77.4 & N/A & N/A\x30& 307 \\
    {BEVT} {\scriptsize{Swin-B}} ~\cite{wang2022bevt} & IN-1K & 81.1 & N/A & 282\x12 & 88 \\
    {MaskFeat} {\scriptsize{MViT-S}} ~\cite{maskfeat} & \multicolumn{1}{c|}{-} & 82.2 & 95.1 & 71\x10 & 36 \\
    VideoMAE {\scriptsize{ViT-S}}~\cite{tong2022videomae} & \multicolumn{1}{c|}{-} & 79.0 & 93.8 & 57\x15 & 22 \\
    VideoMAE {\scriptsize{ViT-B}}~\cite{tong2022videomae} & \multicolumn{1}{c|}{-} & 81.5 & 95.1 & 180\x15 & 87 \\
    VideoMAE {\scriptsize{ViT-L}}~\cite{tong2022videomae} & \multicolumn{1}{c|}{-} & 85.2 & 96.8 & 597\x15 & 305 \\
    VideoMAE {\scriptsize{ViT-H}}~\cite{tong2022videomae} & \multicolumn{1}{c|}{-} & 86.6 & 97.1 & 1192\x15 & 633 \\
    ST-MAE {\scriptsize{ViT-B}}~\cite{feichtenhofer2022masked} & \multicolumn{1}{c|}{-} & 81.3 & 94.9 & 180\x21 & 87 \\
    ST-MAE {\scriptsize{ViT-L}}~\cite{feichtenhofer2022masked} & \multicolumn{1}{c|}{-} & 84.8 & 96.2 & 598\x21 & 304 \\
    ST-MAE {\scriptsize{ViT-H}}~\cite{feichtenhofer2022masked} & \multicolumn{1}{c|}{-} & 85.1 & 96.6 & 1193\x21 & 632 \\
    OmniMAE {\scriptsize{ViT-B}}~\cite{girdhar2022omnimae} & IN-1K & 80.8 & N/A & 180\x15 & 87 \\
    OmniMAE {\scriptsize{ViT-L}}~\cite{girdhar2022omnimae} & IN-1K+SSv2 & 84.0 & N/A & 597\x15 & 305 \\
    OmniMAE {\scriptsize{ViT-H}}~\cite{girdhar2022omnimae} & IN-1K+SSv2 & 84.8 & N/A & 1192\x15 & 633 \\
    \hline
    \textbf{\system-S {\scriptsize{(Teacher-B)}}} & IN-1K & 80.6 & 94.7 & 57\x15 & 22 \\
    \textbf{\system-S {\scriptsize{(Teacher-L)}}} & IN-1K & 81.0  & 94.8  & 57\x15 & 22 \\
    \textbf{\system-B {\scriptsize{(Teacher-B)}}} & IN-1K & 82.7 & 95.4 & 180\x15 & 87 \\
    \textbf{\system-B {\scriptsize{(Teacher-L)}}} & IN-1K & 83.4 & 95.8 & 180\x15 & 87 \\
    \textbf{\system-L {\scriptsize{(Teacher-L)}}} & IN-1K & 86.0 & 96.9 & 597\x15 & 305 \\
    \textbf{\system-L {\scriptsize{(Teacher-L)}}} $\dagger$ & IN-1K & \textbf{86.4} & \textbf{97.0} & 597\x15 & 305 \\
    \textbf{\system-H {\scriptsize{(Teacher-H)}}} $\dagger$ & IN-1K & \textbf{87.2} & \textbf{97.4} & 1192\x15 & 633 \\
\end{tabular}
}
\caption{\textbf{Comparison with previous works on Kinetics-400}. ``MVD-S (Teacher-B)'' denotes that the student is ViT-S and the teacher is ViT-B. $\dagger$ denotes that the student is distilled for 800 epochs instead of 400 epochs. The inference cost is reported with the cost of a single view $\times$ the number of views. Note that IN-1K is only used for training image teachers in MVD.}
\label{tab:k400}
\vspace{-10pt}
\end{table}

\begin{table}[t!]
\centering
\captionsetup[sub]{font=normalsize}
\tablestyle{2.0pt}{1.02}
\resizebox{1.02\linewidth}{!}{
\hspace*{-7pt}
\begin{tabular}{l|l|c|c|c}
    method & extra data & top-1 & \footnotesize{GFLOPs} & \footnotesize{Param} \\
    \shline \hline
    \textit{supervised} & & & & \\
    {SlowFast} R101~\cite{slowfast} & K400 & 63.1  & 106\x3 & 53 \\
    TSM-RGB R50~\cite{tsm} & IN-1K & 63.3 & 62\x6 & 24 \\
    TAM R50~\cite{liu2021tam} & IN-1K & 66.0 & 99\x6 & 51 \\
    {TDN} R101~\cite{wang2021tdn} & IN-1K & 69.6 & 198\x3 & 88 \\
    \hline
    MViTv1-B~\cite{fan2021multiscale} &\multicolumn{1}{c|}{-} & 67.7 & 455\x3 & 37 \\
    MViTv2-B ~\cite{mvitv2} & K400 & 70.5 & 225\x3 & 51 \\
    Uniformer-B~\cite{li2022uniformer} & K400 & 71.2 & 259\x3 & 50 \\
    \hline
    TimeSformer-HR~\cite{timesformer} & IN-21K & 62.5 & 1703\x3 & 121 \\
    ViViT-L FE~\cite{arnab2021vivit} & IN-21K+K400 & 65.9 & 995\x12 & N/A \\
    Mformer-B~\cite{patrick2021keeping} & IN-21K+K400 & 66.5 & 370\x3 & 109 \\
    Mformer-L~\cite{patrick2021keeping} & IN-21K+K400 & 68.1 & 1185\x3 & 382 \\
    VideoSwin-B~\cite{liu2021video} & IN-21K+K400 & 69.6 & 321\x3 & 88 \\
    MViTv2-L ~\cite{mvitv2} & IN-21K+K400 & 73.3 & 2828\x3 & 213 \\
    \shline\hline
    \textit{self-supervised} & & &  & \\
    VIMPAC {\scriptsize{ViT-L}}~\cite{tan2021vimpac} & HowTo100M & 68.1 & N/A\x30 & 307 \\
    {BEVT} {\scriptsize{Swin-B}} ~\cite{wang2022bevt} & IN-1K+K400 & 71.4 & 321\x3 & 88 \\
    {MaskFeat} {\scriptsize{MViT-L}} ~\cite{maskfeat} & K400 & 74.4 & 2828\x3 & 218 \\
    VideoMAE {\scriptsize{ViT-S}}~\cite{tong2022videomae} & K400 & 66.4 & 57\x6 & 22 \\
    VideoMAE {\scriptsize{ViT-S}}~\cite{tong2022videomae} & \multicolumn{1}{c|}{-} & 66.8 & 57\x6 & 22 \\
    VideoMAE {\scriptsize{ViT-B}}~\cite{tong2022videomae} & K400 & 69.7 & 180\x6 & 87 \\
    VideoMAE {\scriptsize{ViT-B}}~\cite{tong2022videomae} & \multicolumn{1}{c|}{-} & 70.8 & 180\x6 & 87 \\
    VideoMAE {\scriptsize{ViT-L}}~\cite{tong2022videomae} & K400 & 74.0 & 597\x6 & 305 \\
    VideoMAE {\scriptsize{ViT-L}}~\cite{tong2022videomae} & \multicolumn{1}{c|}{-} & 74.3 & 597\x6 & 305 \\
    ST-MAE {\scriptsize{ViT-L}}~\cite{feichtenhofer2022masked} & K400 & 72.1 & 598\x3 & 304 \\
    ST-MAE {\scriptsize{ViT-H}}~\cite{feichtenhofer2022masked} & K400 & 74.1 & 1193\x3 & 632 \\
    OmniMAE {\scriptsize{ViT-B}}~\cite{girdhar2022omnimae} & IN-1K & 69.5 & 180\x6 & 87 \\
    OmniMAE {\scriptsize{ViT-B}}~\cite{girdhar2022omnimae} & IN-1K+K400 & 69.0 & 180\x6 & 87 \\
    OmniMAE {\scriptsize{ViT-L}}~\cite{girdhar2022omnimae} & IN-1K & 74.2 & 597\x6 & 305 \\
    OmniMAE {\scriptsize{ViT-H}}~\cite{girdhar2022omnimae} & IN-1K & 75.3 & 1192\x6 & 632 \\
    \hline
    \textbf{\system-S {\scriptsize{(Teacher-B)}}} & IN-1K+K400 & 70.7 & 57\x6 & 22 \\
    \textbf{\system-S {\scriptsize{(Teacher-L)}}} & IN-1K+K400 & 70.9 & 57\x6 & 22 \\
    \textbf{\system-B {\scriptsize{(Teacher-B)}}} & IN-1K+K400 & 72.5 & 180\x6 & 87 \\
    \textbf{\system-B {\scriptsize{(Teacher-L)}}} & IN-1K+K400 & 73.7 & 180\x6 & 87 \\
    \textbf{\system-L {\scriptsize{(Teacher-L)}}} & IN-1K+K400 & 76.1 & 597\x6 & 305 \\
    \textbf{\system-L {\scriptsize{(Teacher-L)}}} $\dagger$ & IN-1K+K400 & \textbf{76.7} & 597\x6 & 305 \\
    \textbf{\system-H {\scriptsize{(Teacher-H)}}} $\dagger$ & IN-1K+K400 & \textbf{77.3} & 1192\x6 & 633 \\
\end{tabular}
}
\caption{\textbf{Comparison with previous works on Something-Something V2.}. $\dagger$ denotes that the student is distilled for 800 epochs.}
\label{tab:ssv2}
\vspace{-8pt}
\end{table}

\vspace{0.05in}
\noindent \textbf{MVD outperforms VideoMAE baseline significantly.}  In Table~\ref{tab:diff_size}, MVD with spatial-temporal co-teaching is compared with VideoMAE
pretrained on K400. When the size of teacher models is the same as that of students, MVD outperforms VideoMAE by a clear margin on both K400 and SSv2. Larger models as teachers can further boost the performance of MVD. It is worth mentioning that not only is our MVD particularly effective for relatively small models, but also it improves the performance of large vision models like ViT-L. For example, with a ViT-L model as the student model, \system achieves 86.0\% and 76.1\% on K400 and SSv2, surpassing the VideoMAE model by 0.8\% and 2.1\%, respectively.

\begin{table}[t!]
\centering
\captionsetup[sub]{font=normalsize}
\tablestyle{2.0pt}{1.02}
\resizebox{1.02\linewidth}{!}{
\hspace*{-7pt}
\begin{tabular}{l|l|c|c|c|c}
    method & extra data & extra labels & mAP & \footnotesize{GFLOPs} & \footnotesize{Param} \\
    \shline \hline
    \textit{supervised} & & & & & \\
    SlowFast R101~\cite{slowfast} & K400 & \cmark & 23.8 & 138 & 53 \\
    MViTv2-B~\cite{mvitv2} & K400 & \cmark & 29.0 & 225 & 51 \\
    MViTv2-L~\cite{mvitv2} & IN-21K+K700 & \cmark & 34.4 & 2828 & 213 \\
    \shline\hline
    \textit{self-supervised} & & & & & \\
    {MaskFeat} {\scriptsize{MViT-L}} ~\cite{maskfeat} & K400 & \cmark & 37.5 & 2828 & 218 \\
    VideoMAE {\scriptsize{ViT-B}}~\cite{tong2022videomae} & K400 & \xmark & 26.7 & 180 & 87 \\
    VideoMAE {\scriptsize{ViT-B}}~\cite{tong2022videomae} & K400 & \cmark & 31.8 & 180 & 87 \\
    VideoMAE {\scriptsize{ViT-L}}~\cite{tong2022videomae} & K400 & \xmark & 34.3 & 597 & 305 \\
    VideoMAE {\scriptsize{ViT-L}}~\cite{tong2022videomae} & K400 & \cmark & 37.0 & 597 & 305 \\
    VideoMAE {\scriptsize{ViT-H}}~\cite{tong2022videomae} & K400 & \xmark & 36.5 & 1192 & 633 \\
    VideoMAE {\scriptsize{ViT-H}}~\cite{tong2022videomae} & K400 & \cmark & 39.5 & 1192 & 633 \\
    ST-MAE {\scriptsize{ViT-L}}~\cite{feichtenhofer2022masked} & K400 & \cmark & 35.7 & 598 & 304 \\
    ST-MAE {\scriptsize{ViT-H}}~\cite{feichtenhofer2022masked} & K400 & \cmark & 36.2 & 1193 & 632 \\
    \hline
    \textbf{\system-B {\scriptsize{(Teacher-B)}}} & IN-1K+K400 & \xmark & 29.3 & 180 & 87 \\
    \textbf{\system-B {\scriptsize{(Teacher-B)}}} & IN-1K+K400 & \cmark & 33.6 & 180 & 87 \\
    \textbf{\system-B {\scriptsize{(Teacher-L)}}} & IN-1K+K400 & \xmark & 31.1 & 180 & 87 \\
    \textbf{\system-B {\scriptsize{(Teacher-L)}}} & IN-1K+K400 & \cmark & 34.2 & 180 & 87 \\
    \textbf{\system-L {\scriptsize{(Teacher-L)}}} & IN-1K+K400 & \xmark & 37.7 & 597 & 305 \\
    \textbf{\system-L {\scriptsize{(Teacher-L)}}} & IN-1K+K400 & \cmark & 38.7 & 597 & 305 \\
    \textbf{\system-H {\scriptsize{(Teacher-H)}}} & IN-1K+K400 & \xmark & \textbf{40.1} & 1192 & 633 \\
    \textbf{\system-H {\scriptsize{(Teacher-H)}}} & IN-1K+K400 & \cmark & \textbf{41.1} & 1192 & 633 \\
\end{tabular}
}
\caption{\textbf{Comparison with previous works on AVA v2.2}. ``Extra labels'' denotes whether the pretrained models are intermediately finetuned on the pretraining video dataset using labels before transferred to AVA. }
\label{tab:ava}
\vspace{-5pt}
\end{table}

\begin{table}[h]
\centering
\captionsetup[sub]{font=normalsize}
\tablestyle{2.0pt}{1.02}
\resizebox{1.02\linewidth}{!}{
\hspace*{-7pt}
\begin{tabular}{l|l|c|c|c}
    method & extra data & Param & UCF101 & HMDB51 \\
    \shline \hline
    VideoMoCo {\scriptsize{R2+1D}}~\cite{pan2021videomoco} & K400 & 15 & 78.7 & 49.2 \\
    MemDPC {\scriptsize{R2D3D}}~\cite{han2020memory} & K400 & 32 & 86.1 & 54.5 \\
    Vi$^2$CLR {\scriptsize{S3D}}~\cite{diba2021vi2clr} & K400 & 9 & 89.1 & 55.7 \\
    CORP {\scriptsize{Slow-R50}}~\cite{hu2021contrast} & K400 & 32 & 93.5 & 68.0 \\
    CVRL {\scriptsize{Slow-R50}}~\cite{cvrl} & K400 & 32 & 92.9 & 67.9 \\
    CVRL {\scriptsize{Slow-R152}}~\cite{cvrl} & K600 & 328 & 94.4 & 70.6 \\
    $\rho$BYOL {\scriptsize{Slow-R50}}~\cite{feichtenhofer2021large} & K400 & 32 & 94.2 & 72.1 \\
    VIMPAC {\scriptsize{ViT-L}}~\cite{tan2021vimpac} & HowTo100M & 307 & 92.7 & 65.9 \\
    VideoMAE {\scriptsize{ViT-B}}~\cite{tong2022videomae} & K400 & 87 & 96.1 & 73.3 \\
    \hline
    \textbf{\system-B {\scriptsize{(Teacher-B)}}} & IN-1K+K400 & 87 & \textbf{97.0}  & \textbf{76.4}  \\
    \textbf{\system-B {\scriptsize{(Teacher-L)}}} & IN-1K+K400 & 87 & \textbf{97.5}  & \textbf{79.7}  \\
\end{tabular}
}
\caption{\textbf{Comparison with previous state-of-the-art methods on UCF101 and HMDB51}.}
\label{tab:ucf_hmdb}
\vspace{-8pt}
\end{table}

\vspace{0.05in}
\noindent \textbf{Comparison with state-of-the-art.}  We compare MVD with prior studies on four video recognition tasks. Results on K400 are shown in Table~\ref{tab:k400}. Our MVD outperforms previous self-supervised methods with similar or less computational cost. Even compared with video transformers pretrained on ImageNet-21K, MVD achieves superior performance. Particularly, \system-H achieves an accuracy of 87.2\% on K400, outperforming previous top-performing methods by clear margins.
Table~\ref{tab:ssv2} presents comparisons to the state-of-the-art methods on SSv2. We observe that for downstream tasks depending on temporal relationship modeling, self-supervised methods based on masked video modeling achieve better performance in comparison with supervised methods (\cf results in the middle group of \Cref{tab:ssv2} \vs results in the top group). 
Once again, our MVD, producing an accuracy of 76.1\% with a large model, beats both supervised methods and self-supervised methods by clear margins. With more training epochs (\ie, 800 epochs), MVD with ViT-L achieves more significant performance gains (\ie, 1.2\%, 2.4\% on K400 and SSv2) compared with VideoMAE. When a huge model is used, the performance can still be boosted and MVD achieves 77.3\% top-1 accuracy on SSv2.

We also evaluate the transfer learning ability of \system on two relatively small datasets, UCF101 and HMDB51. As shown in Table~\ref{tab:ucf_hmdb}, MVD with ViT-B obtains higher accuracy compared with prior works based on well-designed pretext tasks, contrastive learning and masked video modeling methods. Especially compared to the original VideoMAE \cite{tong2022videomae} ViT-B teacher model, we achieve 0.9\% and 3.1\% higher points on these two datasets respectively. Additionally, when teachers with larger size are adopted, MVD achieves stronger transfer learning performance.

When transferred to the more complicated action detection task (AVA v2.2), MVD still shows remarkable improvement compared with previous methods, as shown in Table~\ref{tab:ava}. For example, without additional labels of K400, MVD with ViT-L outperforms VideoMAE by 3.4 to achieve 37.7 mAP. When we intermediately finetune the pretrained models on K400, MVD with ViT-L also achieves significant performance improvement (\ie, 1.7 mAP) compared with VideoMAE. Finally, with a ViT-Huge model, MVD achieves 41.1 mAP, improving 1.6 over the prior state-of-the-art method.

\begin{figure}%
    \centering
    \subfloat[\centering image teacher. ]{\includegraphics[width=0.44\linewidth]{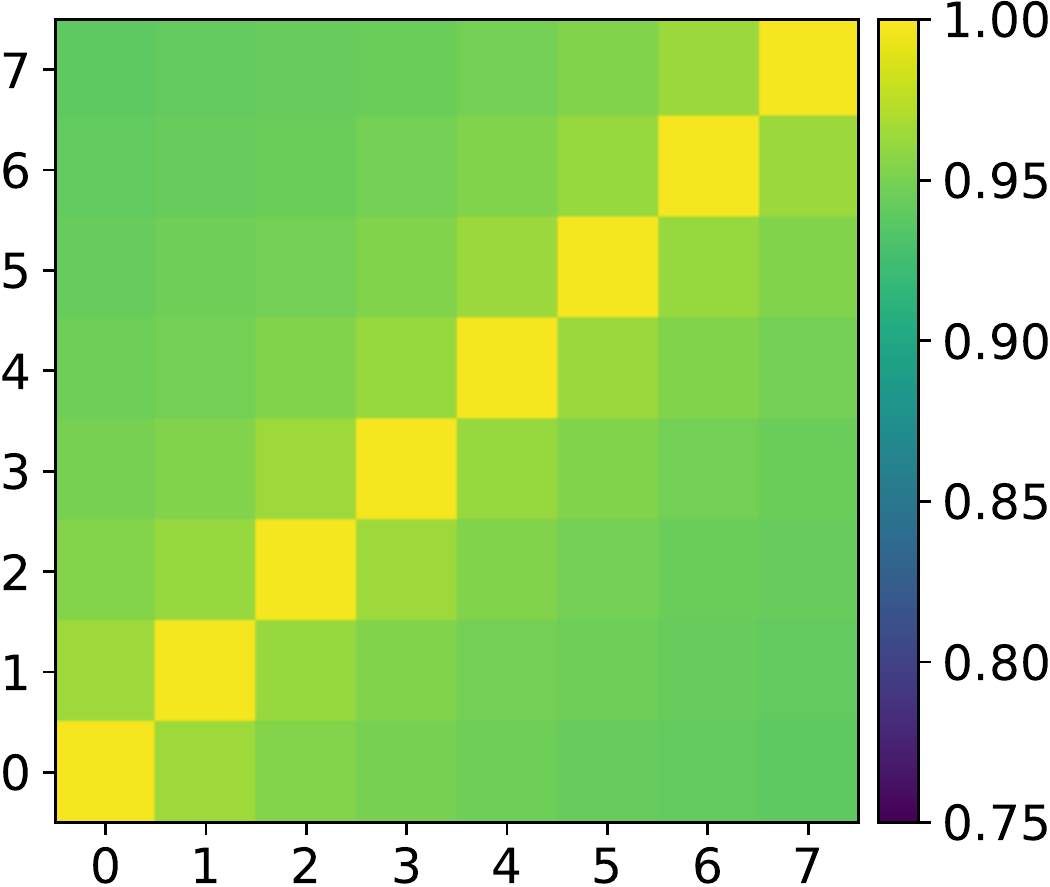} }%
    \qquad
    \subfloat[\centering  video teacher.]{\includegraphics[width=0.44\linewidth]{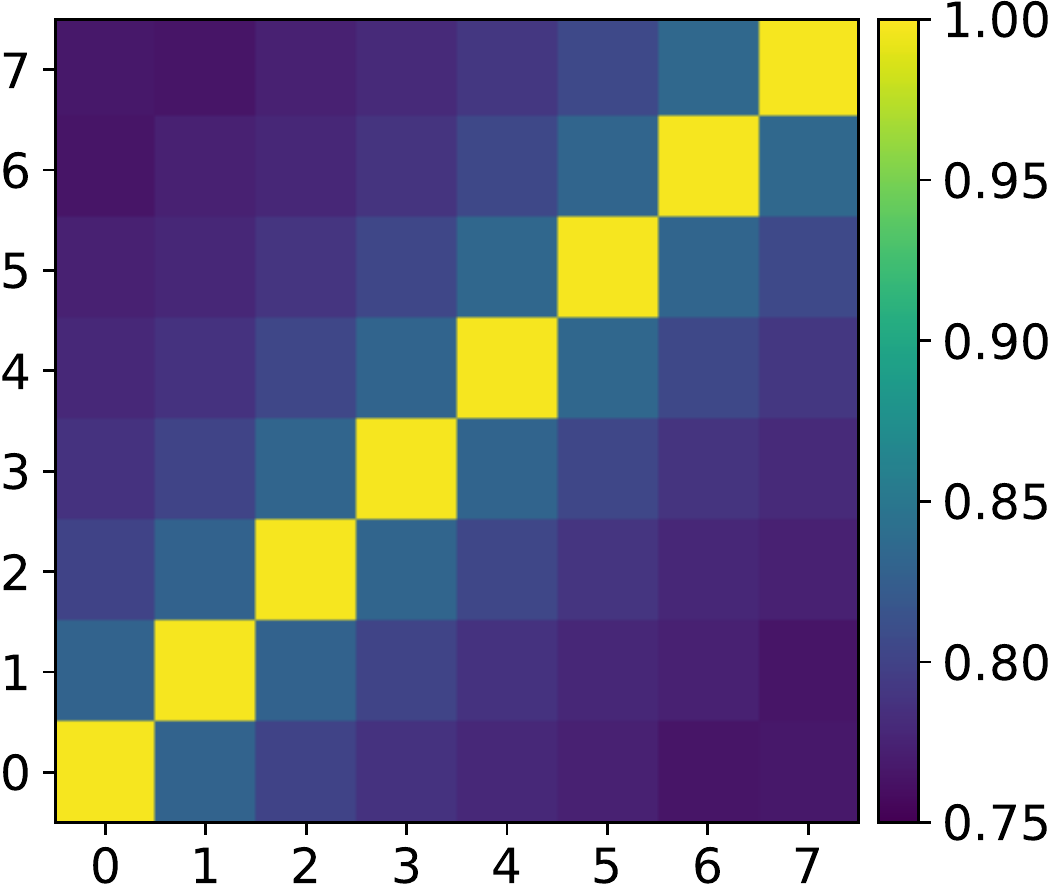} }%
    \caption{\textbf{Feature similarity across different frames for different teacher models.} Similarity matrices are computed on the Kinetics-400 validation set.}%
    \label{fig:teacher_frame_similarity}%
    \vspace{-5pt}
\end{figure}

\begin{table}[t]
\centering
\begin{tabular}{lcc|c}
method & epoch & time(h) & K400 top-1 \\
\shline
VideoMAE & 800 & 107 & 81.0 \\
VideoMAE & 1600 & 214 & 81.5    \\
MVD & (800+) 400 & (107+) 57  & \textbf{81.9} \\
\end{tabular}
\caption{\textbf{Training time comparison between MVD and VideoMAE on 32 NVIDIA V100 GPUs.} The training time of teacher models is considered. we only adopt a video teacher pretrained for 800 epochs in MVD here. ViT-B is used here.}
\label{tab:compare_considering_time}
\vspace{-10pt}
\end{table}

\subsection{Analysis and Discussion}
In this section, we provide an in-depth analysis of the effectiveness of different components in \system. 

\label{sec:ablation}
\noindent \textbf{Analysis of features encoded by different teachers.} The properties of target features generated by different teachers may influence the performance of students on different downstream tasks. To quantify the temporal dynamics that teacher models capture from the input video, we study the similarity between feature maps across different frames of each input video clip via the cosine similarity. As similarity matrices shown in Figure~\ref{fig:teacher_frame_similarity},  for image teachers, the feature maps of different frames are almost the same. However, for video teachers, the features of different frames have larger difference. This indicates that video teachers capture more temporal difference. Therefore, students distilled from video teachers can learn stronger temporal dynamics and perform better on temporally-heavy downstream tasks.

\vspace{0.05in}
\noindent \textbf{Training time comparison.} We study whether MVD is able to achieve better balance of accuracy and efficiency than VideoMAE. For fair comparisons, the training time of teacher models is also counted in the total training time of MVD. Results are shown in Table~\ref{tab:compare_considering_time}. We see that MVD can achieve better accuracy (\ie, 81.9\%) by a total of 164 hours of training which is 50 hours less than 
VideoMAE (producing an accuracy of 81.5\%) trained for 1600 epochs.

\begin{table}[t]
\centering
\begin{tabular}{lc|l}
teachers & reconstruct pixel & SSv2 top-1 \\
\shline
image model & \xmark &  68.7 \\
image model & \cmark &  67.9 \Down{0.8}   \\
\hline
image+video models & \xmark &  70.1 \\
image+video models & \cmark & 69.0 \Down{1.1} \\
\end{tabular}
\caption{\textbf{Ablations on reconstructing low-level features in masked video distillation.} We distill ViT-S students with ViT-B teachers for 300 epochs here.}
\label{tab:low-level-feat}
\vspace{-2pt}
\end{table}

\begin{table}[t]
\centering
\begin{tabular}{l|c|c|c|c}
\multicolumn{1}{c|}{\multirow{2}{*}{teacher}} & \multicolumn{1}{c|}{\multirow{2}{*}{init}} & \multicolumn{1}{c|}{\multirow{2}{*}{epoch}} & \multicolumn{2}{c}{top-1} \\
\cline{4-5} 
\multicolumn{1}{l|}{}     & \multicolumn{1}{l|}{} & \multicolumn{1}{l|}{} & \multicolumn{1}{l|}{K400} & SSv2       \\ 
\shline
momentum encoder & \xmark & 800 & 80.5 & 70.4   \\
momentum encoder & \cmark & 800 & 81.8 & 70.8  \\
fixed image model & \xmark & 400 & 82.3 & 71.4  \\
fixed video model & \xmark & 400 & 82.1 & 71.8   \\
fixed co-teaching & \xmark & 400 & 82.7 & 72.5   \\
\end{tabular}
\caption{\textbf{Comparison with bootstrapped teachers and students initialized with IN1K-pretrained models.} ViT-B is used here.}
\label{tab:bootstrap}
\vspace{-5pt}
\end{table}

\vspace{0.05in}
\noindent \textbf{Reconstruction signals in \system.} In MVD, we first pretrain teacher models by recovering the pixels of masked patches in a MAE manner, then adopt features produced by teacher models as the targets of masked feature modeling. In Table~\ref{tab:low-level-feat}, we study whether to include an additional decoder branch for reconstructing pixels of masked patches in the distillation stage. The experimental results show that for both MVD with a single teacher and MVD with spatial-temporal co-teaching, the reconstruction of low-level feature targets degrades the performance for downstream tasks. Therefore, we only reconstruct high-level features of masked patches in the distillation stage of MVD.

\vspace{0.05in}
\noindent \textbf{Comparison with bootstrapped teachers.} Some recent approaches of image representation learning~\cite{zhou2021ibot,bootmae,chen2022sdae,dong2022maskclip} adopt features of a momentum encoder as the targets of masked image modeling, while we use frozen teacher models in MVD. In Table~\ref{tab:bootstrap}, we compare fixed teachers with bootstrapped teachers that are updated by an exponential moving average of the online encoder during pretraining. According to a masked image modeling method~\cite{bootmae}, two strong baselines of bootstrapped teachers are built for video representation learning: (a) The student model is trained from scratch with masked feature modeling, and the target features are generated by a momentum encoder. (b) The framework of bootstrapped teachers is first pretrained on IN-1K for 800 epochs, then the pretrained weights are adopted to initialize the video pretraining. As shown in Table~\ref{tab:bootstrap}, MVD with the frozen teacher beats the method with a bootstrapped teacher on the downstream video tasks, even if only a single teacher is utilized. 

\begin{table}[t]
\centering
\begin{tabular}{l|c|c}
distillation method &  K400 top-1 &  SSv2 top-1  \\
\shline
per-token distillation & 80.9 &  70.5  \\
masked reconstruction & 82.1  & 71.8  \\

\end{tabular}
\caption{\textbf{Comparison with feature distillation without masked reconstruction.} We distill ViT-B models from the video teacher here.}
\label{tab:feat_distill}
\vspace{-5pt}
\end{table}

\vspace{0.05in}
\noindent \textbf{Comparison with feature distillation.} In the previous work of self-supervised feature distillation~\cite{fang2021seed,gou2021knowledge,xu2021bag}, the distillation loss is directly computed upon the full feature maps between teachers and students. Accordingly, we build a baseline method named per-token distillation. Specifically, the output features of students are projected by a MLP, and then forced to mimicking the teacher's feature at each token with a Smooth L1 loss. As shown in Table~\ref{tab:feat_distill}, masked feature reconstruction in our MVD outperforms per-token distillation on both K400 and SSv2.

\section{Conclusion}
In this paper, we study masked video distillation upon MIM pretrained image or video transformers. We have three interesting findings: 1) Using MIM pretrained image transformers or MVM pretrained video transformers as teachers supervise masked feature prediction can significantly boost the finetuning performance on video downstream tasks; 2) The representation distilled with image and video teachers will have different properties, i.e., image teachers will benefit spatial-heavy video tasks more while video teachers benefit temporal-heavy video tasks more; 3) Combining image and video teachers will enjoy the synergy and thus produce higher performance. Even though the proposed masked video distillation seems very straightforward, we hope such interesting findings can motivate more thinking about masked video pretraining.

{\small
\bibliographystyle{ieee_fullname}
\bibliography{egbib}
}

\clearpage
\newpage
\appendix

\section{Additional Results}

\subsection{Masked Feature Modeling for Image Models}

We perform masked reconstruction of high-level features for the image ViT on ImageNet-1K. For masked feature modeling on the image data, only the image teacher in MVD can be used. As the results shown in Table~\ref{tab:imagenet_vs_video}, compared with the MAE baseline, masked feature distillation achieves 0.4\% Top-1 accuracy gain on ImageNet-1K. When comparing the performance improvement against masked reconstruction of pixels between image models and video models, we observe that MVD achieves greater performance gains on video downstream tasks.  

\begin{table}[h!]
\centering
\begin{tabular}{l|c|c|c|c}
\multicolumn{1}{c|}{\multirow{2}{*}{target}} & \multicolumn{1}{c|}{\multirow{2}{*}{epoch}} & \multicolumn{3}{c}{top-1 accuracy} \\
\cline{3-5} 
\multicolumn{1}{l|}{} & \multicolumn{1}{l|}{} & \multicolumn{1}{c|}{IN-1K} & \multicolumn{1}{c|}{K400}  & SSv2       \\ 
\shline
pixels & 1600 & 83.6 & 81.5 & 69.7   \\
features & 400 & 84.0\Up{0.4} & 82.7\Up{1.2} & 72.5\Up{2.8}  \\
\end{tabular}
\caption{\textbf{Comparison with masked feature modeling for image models.} We distill ViT-B for 400 epochs here.}
\label{tab:imagenet_vs_video}
\end{table}

\section{Implementation Details}

\subsection{Pretraining Experiments}

We pretrain image teacher models on ImageNet-1K following the strategy in ~\cite{he2021masked}, and pretrain video teacher models on Kinetics-400 following the strategy in ~\cite{tong2022videomae}. For the distillation stage in MVD, we distill student models with teacher models for 400 epochs on Kinetics-400 unless otherwise stated. The length of input videos is 16 frames during pretraining. We adopt tube masking in ~\cite{tong2022videomae} and the masking ratio in the distillation stage is 90\%. The default setting of pretraining is presented in Table~\ref{tab:pretraining_setting}.

\begin{table}[h]
\centering
\begin{tabular}{l|c}
config & Kinetics-400 \\
\shline
optimizer & AdamW~\cite{adamw}   \\
base learning rate & 1.5e-4    \\
weight decay & 0.05 \\
optimizer momentum & $\beta_1$,$\beta_2$=0.9,0.95~\cite{chen2020generative} \\
batch size & 1024 (S,B), 512 (L) \\
learning rate schedule & cosine decay~\cite{coslr} \\
warmup epochs & 40 \\
augmentation & MultiScaleCrop~\cite{tsn} \\
drop path & 0.1 (S,B), 0.2(L) \\
\end{tabular}
\caption{\textbf{Pretraining setting of MVD.}}
\label{tab:pretraining_setting}
\end{table}

\begin{table*}[h]
\centering
\begin{tabular}{l|cccc}
config & Sth-Sth V2 & Kinetics-400 & UCF101 & HMDB51 \\
\shline
optimizer & \multicolumn{4}{c}{AdamW} \\
base learning rate & 1e-3(S), 5e-4(B,L) & 1e-3 & 5e-4 & 1e-3 \\
weight decay & \multicolumn{4}{c}{0.05} \\
optimizer momentum & \multicolumn{4}{c}{$\beta_1, \beta_2{=}0.9, 0.999$} \\
batch size & 512 & 512 & 128 & 128\\
learning rate schedule & \multicolumn{4}{c}{cosine decay} \\
warmup epochs & \multicolumn{4}{c}{5} \\
training epochs &  40~(S), 30~(B,L) & 150~(S), 75~(B), 50~(L) & 100 & 50 \\
repeated augmentation & \multicolumn{4}{c}{2}  \\
flip augmentation & \emph{no} & \emph{yes} & \emph{yes} & \emph{yes} \\
RandAug~\cite{cubuk2020randaugment}  & \multicolumn{4}{c}{(9, 0.5)} \\
label smoothing~\cite{label_smoothing}  & \multicolumn{4}{c}{0.1} \\
mixup~\cite{zhang2017mixup}  & \multicolumn{4}{c}{0.8}  \\
cutmix~\cite{yun2019cutmix}  & \multicolumn{4}{c}{1.0} \\
drop path~\cite{droppath} & \multicolumn{2}{c}{0.1~(S,B), 0.2~(L)} & 0.2 & 0.2 \\
dropout~\cite{hinton2012improving_dropout} & 0.5~(L) & 0.5~(L) & 0.5 & 0.5 \\
layer-wise lr decay~\cite{bao2021beit}  & 0.7~(S),0.75~(B,L) & 0.75 & 0.7 & 0.7 \\
\end{tabular}
\caption{\textbf{Fine-tuning setting of MVD.}}
\label{tab:ft_setting}
\end{table*}

\subsection{Finetuning Experiments}

We transfer models pretrained by MVD on Kinetics-400 to video downstream tasks with the default setting in Table~\ref{tab:ft_setting}.

\vspace{0.05in}
\noindent \textbf{Kinetics experiments.} When finetuning on Kinetics-400, we adopt the dense sampling following~\cite{quovadis,slowfast} and the default length of input videos is 16 frames. For inference, we use 3 spatial crops \x~5 temporal clips.

\vspace{0.05in}
\noindent \textbf{Something-Something v2 experiments.} During finetuning on Something-Something v2, we adopt the uniform sampling following~\cite{tsn} and the default length of input videos is 16 frames. For inference, we use 3 spatial crops \x~2 temporal clips.

\vspace{0.05in}
\noindent \textbf{UCF101 and HMDB51 experiments.} For finetuning on UCF101 and HMDB51, we adopt the dense sampling and the default length of input videos is 16 frames. For inference, we use 3 spatial crops \x~5 temporal clips.

\vspace{0.05in}
\noindent \textbf{AVA experiments.} When finetuning on AVA v2.2, following~\cite{tong2022videomae}, we adopt the detection architecture in ~\cite{slowfast} and the detected person boxes from AIA~\cite{tang2020asynchronous}. The default length of input videos is 16 frames. We also use the default finetuning setting in \cite{tong2022videomae} for a fair comparison.

\section{Visualization}

\subsection{Analysis of temporal dynamics}

In our paper, to quantify the temporal dynamics that models capture from the input video, we study the similarity between feature maps across different frames of each input video clip via the cosine similarity. 

\begin{figure*}%
    \centering
    \subfloat[\centering image teacher. ]{\includegraphics[width=0.35\linewidth]{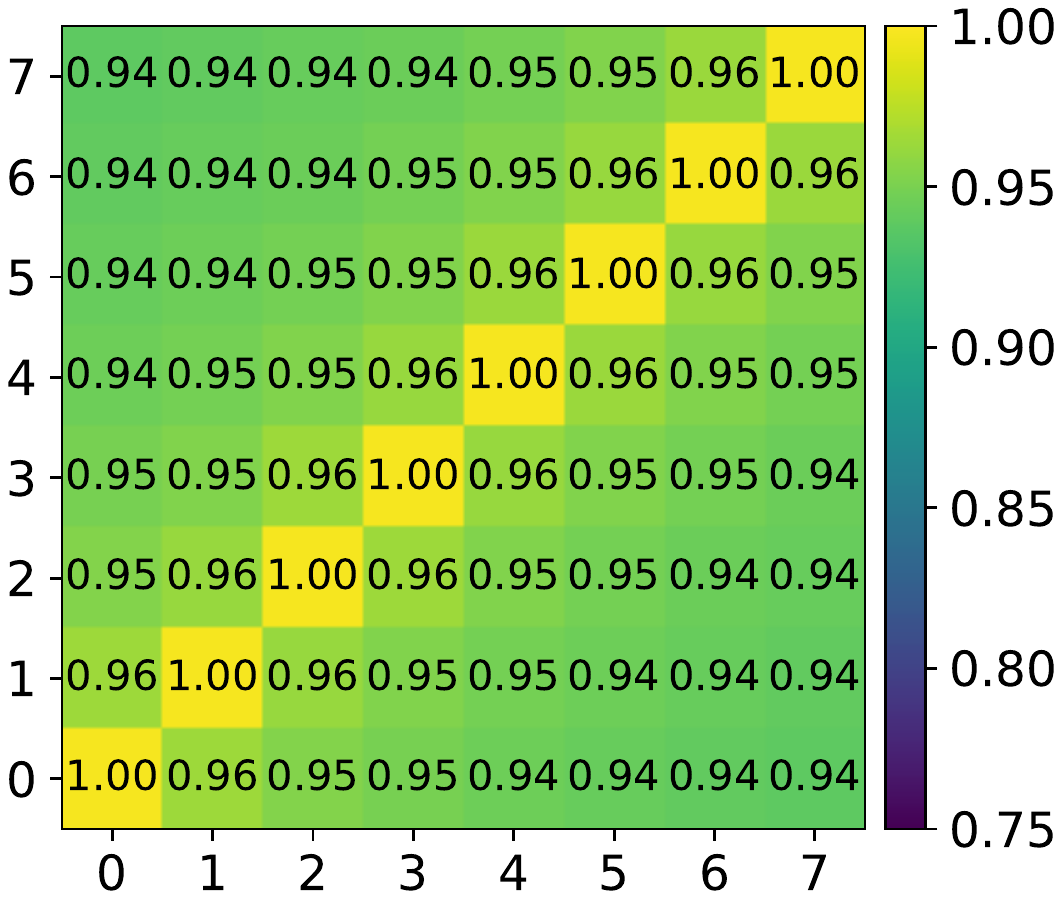} }%
    \qquad
    \subfloat[\centering  video teacher.]{\includegraphics[width=0.35\linewidth]{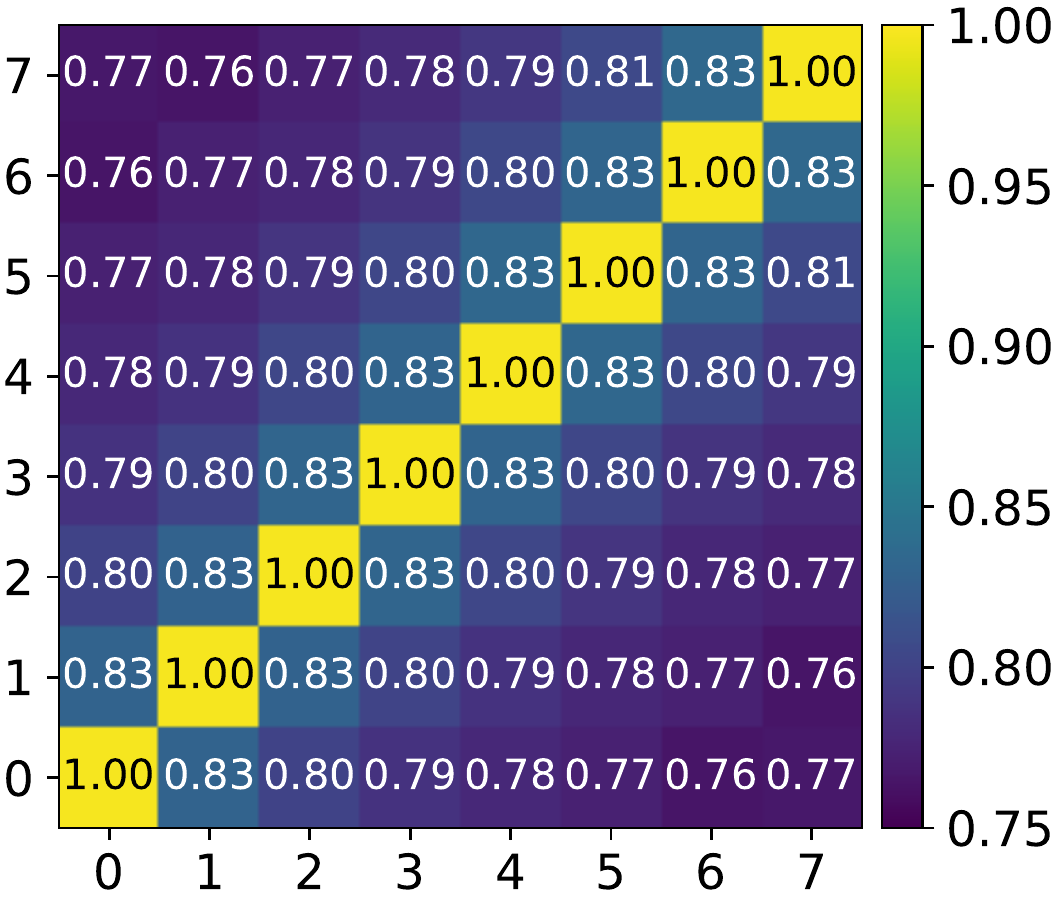} }%
    \caption{\textbf{Feature similarity across different frames for different teacher models.} Similarity matrices are computed on the Kinetics-400 validation set. The numbers in the grid are the values of cosine similarity between two frame features. }%
    \label{fig:teacher_frame_similarity_with_num}%
\end{figure*}

\noindent \textbf{Analysis of features encoded by different teachers.}
The properties of target features generated by different teachers may influence the performance of students on different downstream tasks. As similarity matrices shown in Figure~\ref{fig:teacher_frame_similarity_with_num},  for image teachers, the feature maps of different frames are almost the same. However, for video teachers, the features of different frames have larger differences. This indicates that video teachers capture more temporal difference. Therefore, students distilled from video teachers can learn stronger temporal dynamics and perform better on temporally-heavy downstream tasks.

\begin{figure*}%
    \centering
    \subfloat[\centering student distilled from the image teacher. ]{\includegraphics[width=0.35\linewidth]{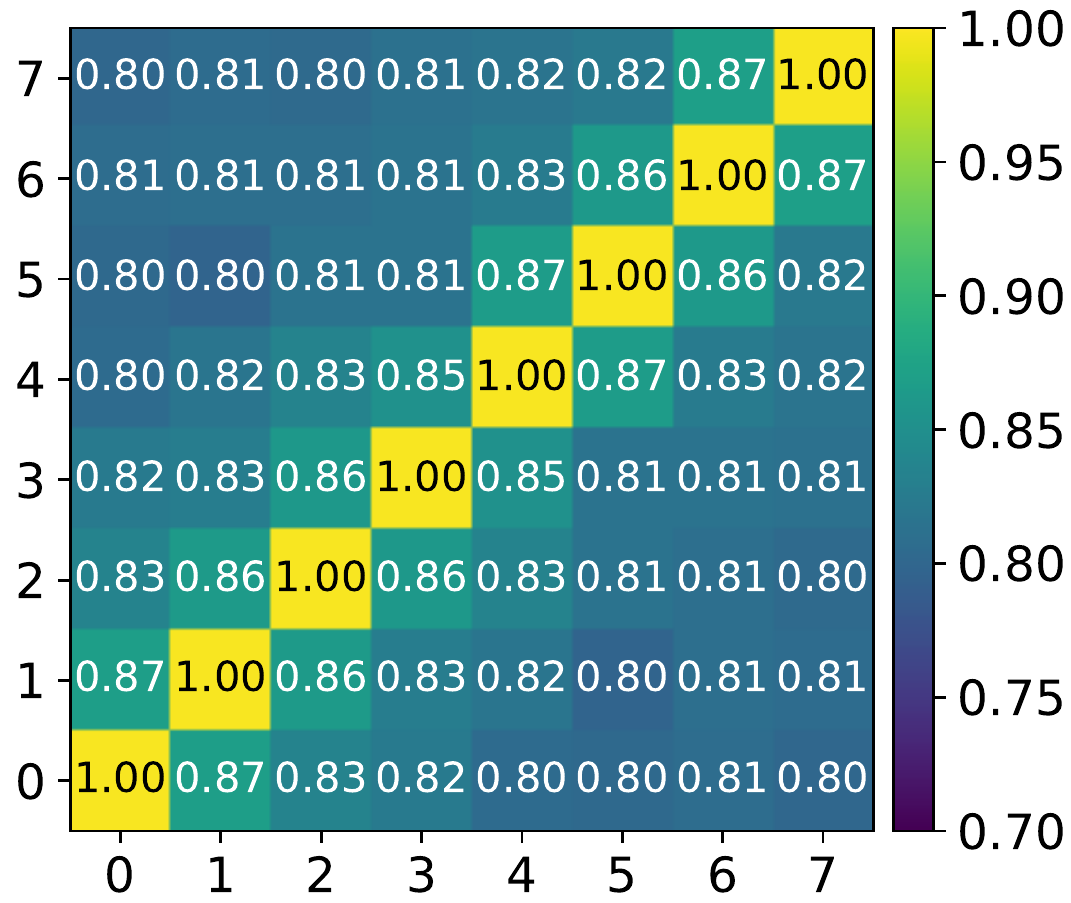} }%
    \qquad
    \subfloat[\centering  student distilled from the video teacher.]{\includegraphics[width=0.35\linewidth]{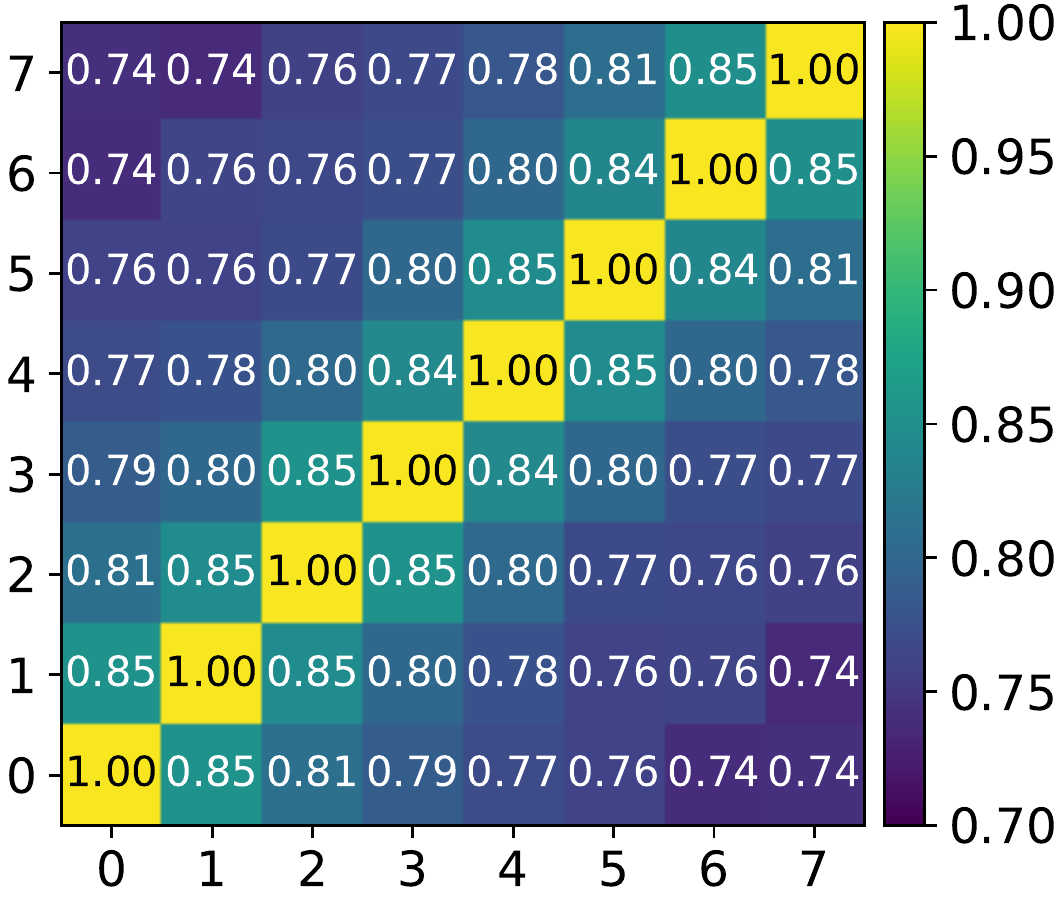} }%
    \caption{\textbf{Feature similarity across different frames for student models distilled from different teacher models.} Similarity matrices are computed on the Kinetics-400 validation set.}%
    \label{fig:student_frame_similarity_with_num}%
\end{figure*}

\noindent \textbf{Analysis of features encoded by students distilled from different teachers.} To study what students learn from different teachers, we visualize the feature similarity across different frames for student models. As results shown in Figure~\ref{fig:student_frame_similarity_with_num}, we observe that (a) for the student distilled from the image teacher, the features of different frames have larger differences compared with those encoded by the image teacher. This indicates that students can learn temporal dynamics from the masked reconstruction of spatial features on videos. (b) For the student distilled from the video teacher, the features of different frames have larger differences compared with those encoded by the student distilled from the image teacher. This demonstrates that students learn stronger temporal dynamics from video teachers.

\end{document}